\documentclass[11pt]{article}
\usepackage{eacl2017}
\usepackage{times}
\usepackage{url}
\eaclfinalcopy
\usepackage{latexsym}
\usepackage{array}
\usepackage{graphicx}
\usepackage{graphics}
\newcolumntype{?}[1]{!{\vrule width #1}}
\usepackage{booktabs}
\usepackage[utf8]{inputenc}
\usepackage{multirow}
\usepackage[normalem]{ulem}
\usepackage[stfloats]{}
\usepackage{multirow}

\title{Automatic Identification of Sarcasm Target: An Introductory Approach}
\author{\begin{tabular}{ccc}
Aditya Joshi$^{2,3,4}$ & &Pranav Goel$^{1}$\\
Pushpak Bhattacharyya$^{2}$&&Mark James Carman$^{3}$\\
\end{tabular}\\
\begin{tabular}{ccc}
 \multicolumn{3}{c}{$^{1}$IIT-BHU (Varanasi), India, $^{2}$IIT Bombay, India, $^{3}$Monash University, Australia}\\
\multicolumn{3}{c}{$^{4}$IITB-Monash Research Academy, India}\\
\multicolumn{3}{c}{\tt \ pranav.goel.cse14@iitbhu.ac.in}, \tt \{adityaj, pb\}@cse.iitb.ac.in\\
\multicolumn{3}{c}{\tt mark.carman@monash.edu}\\
\end{tabular}
}
\begin{document}
\maketitle
\begin{abstract}
Past work in computational sarcasm deals primarily with sarcasm detection. In this paper, we introduce a novel, related problem: sarcasm target identification (\textit{i.e.}, extracting the target of ridicule in a sarcastic sentence). We present an introductory
approach for sarcasm target identification. Our approach employs two types of extractors: one based on rules, and another consisting of a statistical classifier. To compare our approach, we use two baselines: a na\"{i}ve baseline and another baseline based on work in sentiment target identification. We perform our experiments on book snippets and tweets, and show that our hybrid approach performs better than the two baselines and also, in comparison with using the two extractors individually. Our introductory approach establishes the viability of sarcasm target identification, and will serve as a baseline for future work.
\end{abstract}
\textit{This paper was uploaded to arXiv on 20 October, 2016; but was submitted to EACL 2017 at an earlier date. The paper was not on arXiv at the time of EACL submission.}
\section{Introduction}
Sarcasm is a form of verbal irony that is intended to express contempt or ridicule\footnote{Source: The Free Dictionary}. Past work in computational sarcasm deals primarily with sarcasm detection, \textit{i.e.}, to predict whether or a not a given piece of text is sarcastic~\cite{joshi2016automatic}. So the sentence `\textit{A woman needs a man like fish needs bicycle\footnote{This quote is attributed to Irina Dunn, an Australian writer and social activist.}}' will be predicted as sarcastic. While several approaches have been reported for sarcasm detection~\cite{3,4,6,22}, no past work, to the best of our knowledge, attempts to identify a crucial component of sarcasm: the target of ridicule~\cite{campbell2012there}. In case of the example above, this target of ridicule is the word `\textit{man}'.  %Therefore, it is useful that the correct target of ridicule in a sarcastic text be identified. %Towards this, in this paper, we introduce a novel problem: `sarcasm target identification'. The goal is to identify the target of ridicule in a sarcastic text. . Thus, in the woman-man example above, the sarcasm target would be identified as the word `man'.

In this paper, we introduce a new avenue in computational sarcasm research. We explore a novel problem called `\textbf{sarcasm target identification}': the task of extracting the target of ridicule (\textit{i.e.}, \textbf{sarcasm target}) of a sarcastic text. This sarcasm target is either a subset of words in the sentence or a fallback label `Outside'\footnote{This label is necessary because the sarcasm target may not be present as a word in the sentence. Section 2 discusses this in detail.}. We present an introductory approach \textbf{that takes as input a sarcastic text and returns its sarcasm target}. Our hybrid approach employs two extractors: a rule-based extractor (that implements a set of rules) and a statistical extractor (that uses a word-level classifier for every word in the sentence, to predict if the word will constitute the sarcasm target). We evaluate our approach using two manually labeled datasets consisting of book snippets and tweets. We consider two versions of our approach: Hybrid OR (where prediction by the two extractors is OR-ed) and Hybrid AND (where prediction by the two extractors is AND-ed). Since this is the first work in sarcasm target detection, no past work exists to be used as a baseline. Hence, we devise two baselines to validate the strength of our work. The first is a simple, intuitive baseline to show if our approach (which is computationally more intensive than the simple baseline) holds any value\footnote{In absence of past work, using simple and obvious techniques to solve a problem have been considered as baselines in sentiment analysis~\cite{tan2011user,pang2005seeing}}. As the second baseline, we use a technique reported for sentiment/opinion target identification. For both our datasets, we observe that the hybrid approach outperforms both the baselines. In addition, the hybrid OR approach also works better than using either rule-based or statistical extractors individually.

Sarcasm target identification will be useful for aspect-based sentiment analysis so that the negative sentiment expressed in the sarcastic text can be attributed to the correct aspect. To the best of our knowledge, this is the first work that attempts identification of sarcasm targets. Our results will serve as a baseline for future work. Our manually labeled datasets are available for download at: \url{Anonymous}. Each unit consists of a piece of text (either book snippet or tweet) with the annotation as the sarcasm target.
%While past approaches in computational sarcasm deal with identifying whether or not a given text is sarcastic, 

 %However, like aspect-specific sentiment analysis (where, in addition to sentiment polarity, the target of sentiment needs to be attributed correctly), identification of targets of sarcasm 

%Sarcasm detection aims to predict if a given piece of text is sarcastic. Several approaches to sarcasm detection have been proposed. \cite{?} show that sarcasm detection cascaded with sentiment analysis helps sentiment analysis.

%While several sarcasm detection approaches have been reported, no work, to the best of knowledge, attempts to

 %However, like aspect-specific sentiment analysis (where,  identify target of sarcasm.

%We name the task `sarcasm target identification'. The goal of sarcasm target identification is to identify which words in a sarcastic text refer to the target of ridicule. This also means that our datasets consist only of sarcastic text. This means that after a sentence has been detected as sarcastic, our approach can be used to attribute the negative sentiment to the right target.

The rest of the paper is organized as follows. Section~\ref{sec:motiv} formulates the problem, while Section~\ref{sec:archi} describes our architecture in detail. Experiment setup is in Section~\ref{sec:expsetup}. The results are presented in Section~\ref{sec:results} while an error analysis is in Section~\ref{sec:discuss}. We present  related work in Section~\ref{sec:relwork} and conclude the paper in Section~\ref{sec:concl}.
\begin{figure}[ht!]
\centering
        \includegraphics[width=0.53\textwidth]{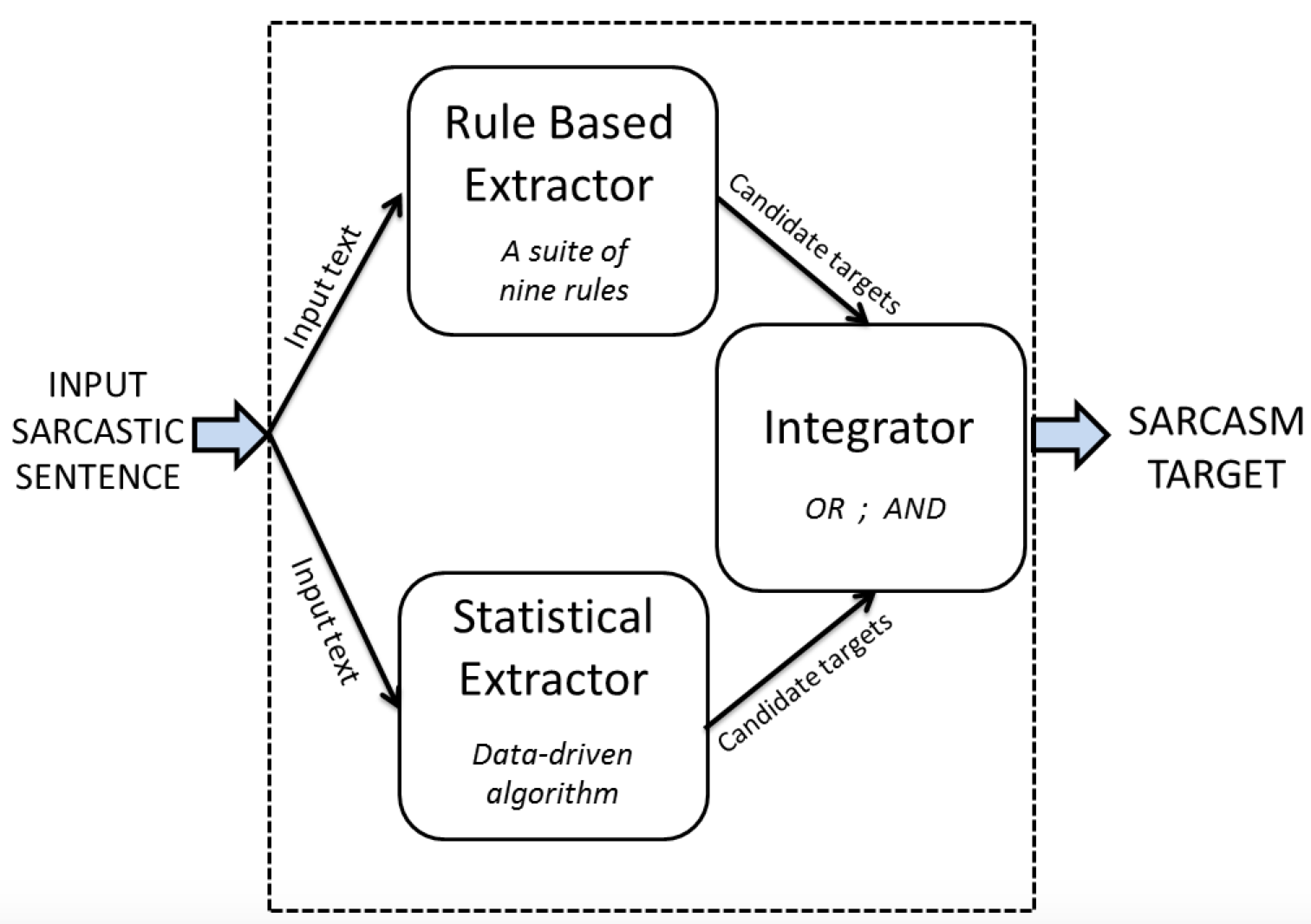}
   \caption{Architecture of our Sarcasm Target Identification Approach}
   \label{fig:archi}
   \end{figure}
\section{Formulation}
\label{sec:motiv}
Sarcasm is a well-known challenge to sentiment analysis~\cite{pang2008opinion}. Consider the sarcastic sentence `\textit{My cell phone has an awesome battery that lasts 20 minutes}'. This sentence ridicules the battery of the cell phone. Aspect-based sentiment analysis needs to identify that the sentence ridicules the battery of the phone and hence, expresses negative sentiment towards the aspect `\textit{battery}'. Our proposed problem `sarcasm target identification' thus enables aspect-based sentiment analysis to attribute the negative sentiment to the correct target aspect. We define the \textbf{sarcasm target} as the entity or situation being ridiculed in a sarcastic text. In case of `\textit{Can't wait to go to class today}', the word `\textit{class}' is the sarcasm target. Every sarcastic text has at least one sarcasm target (by definition of sarcasm), and the notion of sarcasm target is applicable for only sarcastic text (\textit{i.e.}, non-sarcastic text does not have sarcasm target). Thus, we define \textbf{sarcasm target identification} as the task of extracting the subset of words that indicate the target of ridicule, given a sarcastic text. In case the target of ridicule is not present among these words, a fallback label `Outside' is expected. Examples of some sarcasm targets are given in Table~\ref{tab:examples}. %As an example of sarcasm target, consider the sarcastic sentence. 

Some challenges of sarcasm target identification are:
\begin{itemize}\setlength\itemsep{0cm}
\item \textbf{Presence of multiple candidate phrases}: Consider the sentence `\textit{This phone heats up so much that I strongly recommend chefs around the world to use it as a cook-top}'. In this sentence, the words `\textit{chefs}', `\textit{cook-top}' and `\textit{phone}' are candidate phrases. However, only the `\textit{phone}' is being ridiculed in this sentence. 
\item \textbf{Multiple sarcasm targets}: A sentence like `\textit{You are as good at coding as he is at cooking}' ridicules both `\textit{you}' and `\textit{he}', and hence, both are sarcasm targets.
\item \textbf{Absence of a sarcasm target word (the `Outside' case)}: Consider the situation where a student is caught copying in a test, and the teacher says, `\textit{Your parents must be so proud today!}'. No specific word in the sentence is the sarcasm target. The target here is the student. We refer to such cases as the `outside' cases. 
\end{itemize}
       \begin{table*}[h!]
\centering \small
\begin{tabular}{p{13cm}l}
\toprule
\textbf{Example} & \textbf{Target}     \\ \midrule
Love when you don't have two minutes to send me a quick text. & you \\
Don't you just love it when Microsoft tells you that you're spelling your own name wrong. & Microsoft \\
I love being ignored. & being ignored \\%                     Oh I so love this jacket. & this jacket \\
He is as good at coding as Tiger Woods is at avoiding controversy.& He, Tiger Woods \\
Yeah, right! I hate catching the bus on time anyway! & Outside   \\\bottomrule
\end{tabular}
\caption{Examples of sarcasm targets}
\label{tab:examples}
\end{table*}

\section{Architecture}
\label{sec:archi}
%In this section, we first define the notions of sarcasm target and sarcasm target identification, and then introduce the architecture of our approach for sarcasm target identification.
 %We only look at the words within the text as potential candidates for the sarcasm target. If the target lies outside the text, we assign the special label 'OUTSIDE' to that instance. Our hybrid approach aims to automate the task in accordance with this definition.
Our hybrid approach for sarcasm target identification is depicted in Figure~\ref{fig:archi}. The input is a sarcastic sentence while the output is the sarcasm target. The approach consists of two kinds of extractors: (a) a \textbf{rule-based extractor} that implements nine rules to identify different kinds of sarcasm targets, and (b) a \textbf{statistical extractor} that uses statistical classification techniques. The two extractors individually generate lists of candidate sarcasm targets. The third component is the \textbf{integrator} that makes an overall prediction of the sarcasm target by choosing among the sarcasm targets returned by the individual extractors. The overall output is a subset of words in the sentence. In case no word is found to be a sarcasm target, a fallback label `Outside' is returned. In the forthcoming subsections, we describe the three modules in detail.

%\section{Approach}
%We used 4 approaches for automatic identification of sarcasm target : \textbf{rule-based}, \textbf{Statistical}, \textbf{Hybrid} and \textbf{Sequence Labeling}. These have been described in detail below:
\subsection{Rule-based Extractor}
Our rule-based extractor consists of \textbf{nine rules} that 
take as input the sarcastic sentence, and return a set of candidate sarcasm targets. The rules are summarized in Table~\ref{tab:rules}. We now describe each rule, citing past work that motivated the rule, wherever applicable:

%These rules are based on the major patterns and properties we observe throughout our dataset. We will briefly explain all the rules first and then explain the working of this approach i.e how these rules are applied to return their target, and the way we combine the individual rules' results to predict the target. The strength of all these rules and the accuracy of this approach will be discussed in the Results section.
\begin{enumerate}\setlength\itemsep{0cm}
\item \textbf{R1} (\textbf{Pronouns and Pronominal Adjectives}): R1 returns pronouns such as `\textit{you, she, they}' and pronominal adjectives (followed by their object) (as in the case of~`\textit{your shoes}'). Thus, for the sentence `\textit{I am so in love with my job}', the phrases `\textit{I}' (pronoun) and `\textit{my job}' (based on the pronominal adjective `my') are returned as candidate sarcasm targets. This is based on observations by ~\newcite{shamay2005neuroanatomical}.
\item \textbf{R2} (\textbf{Named Entities}): Named entities in a sentence may be sarcasm targets. This rule returns all \textit{named entities} in the sentence. In case of `\textit{Olly Riley is so original with his tweets}', R2 predicts the phrase `\textit{Olly Riley}' as a candidate sarcasm target.
 \item \textbf{R3} (\textbf{Sentiment-bearing verb as the pivot}): This rule is based on the idea by \newcite{riloff} that sarcasm may be expressed as a contrast between a positive sentiment verb and a negative situation. In case of `\textit{I love being ignored}', the sentiment-bearing verb `\textit{love}' is positive. The object of `\textit{love}' is `\textit{being ignored}'. Therefore, R3 returns `\textit{being ignored}' as the candidate sarcasm target. If the sentiment-bearing verb is negative, the rule returns `\textit{Outside}' as a candidate sarcasm target. This is applicable in case of situations like humble bragging\footnote{\url{http://www.urbandictionary.com/define.php?term=humblebrag}} as in `\textit{I hate being so popular}' where the speaker is either ridiculing the listener or just bragging about themselves. 
      \item \textbf{R4} (\textbf{Non-sentiment-bearing verb as the pivot}): This rule applies in case of sentences where the verb does not bear sentiment. The rule identifies which out of subject or object has a lower sentiment score, and returns the corresponding portion as the candidate sarcasm target. For example, rule R4 returns `\textit{to have a test on my birthday}' as the candidate sarcasm target in case of `\textit{Excited that the teacher has decided to have a test on my birthday!}' where `decided' is the non-sentiment-bearing verb. This is also based on ~\newcite{riloff}.
      \item \textbf{R5} (\textbf{Gerundial verb phrases and Infinitives}): R5 returns the gerundial phrase `\textit{being covered in rashes}' in case of `\textit{Being covered in rashes is fun.}' as the candidate sarcasm target. Similarly, in case of `\textit{Can't wait to wake up early to babysit!}', the infinitive `\textit{to wake up early to babysit}' is returned.
    
   \item \textbf{R6} (\textbf{Noun phrases containing positive adjective}): R6 extracts noun phrases of the form `JJ NN' where JJ is a positive adjective, and returns the noun indicated by NN. Specifically, 1-3 words preceding the nouns in the sentence are checked for positive sentiment. In case of `\textit{Look at the most realistic walls in a video game}', the noun `\textit{walls}' is returned as the sarcasm target.
       \item\textbf{R7} : \textbf{Interrogative sentences}: R7 returns the subject of an interrogative sentence as the sarcasm target. Thus, for `\textit{A murderer is stalking me. Could life be more fun?}', the rule returns `\textit{life}' as the target. %If sentence has `\textit{this}' instead of a noun as its subject, the rule returns the object as the target.

       \item \textbf{R8} : \textbf{Sarcasm in Similes}: This rule captures the \textit{subjects/noun phrases} involved in similes and `as if' comparisons. The rule returns the subject on both sides, as in `\textit{He is as good at coding as Tiger Woods is at avoiding controversy.}' Both `\textit{He}' and `\textit{Tiger Woods}' are returned as targets. This is derived from work on sarcastic similes by ~\newcite{veale2010detecting}.
    
       \item \textbf{R9} : \textbf{Demonstrative adjectives}: This rule captures nouns associated with demonstrative adjectives - \textit{this/that/these/those}. For example, for the sentence `\textit{Oh, I love this jacket!}', R9 returns `\textit{this jacket}' as the sarcasm target.
\end{enumerate}
     \begin{table*}[]
     \centering
\small
 \begin{tabular}{p{0.5cm}p{6cm}p{8cm}}
\toprule
 \textbf{Rule} & \textbf{Definition}                                                                                                              & \textbf{Example}                                                                                                                                                                                 \\ \midrule
 R1   & Return pronouns (inluding possessive) and  pronoun based adjectives  & Love when \textbf{you} don't have two minutes to send me a quick text .. ;  I am so in love with \textbf{my job}.                                           \\ 
 R2   & Return named entities as target & Don't you just love it when \textbf{Microsoft} tells  you that you're spelling your own name wrong.                                                 \\ 
 R3   & Return direct object of a positive sentiment  verb                 & I \textit{love} \textbf{being ignored}.                                                                                                                                                                   \\ 
 R4   & Return phrase on lower sentiment side of  primary verb             & So happy to just find out it has been \textit{decided} \textbf{to reschedule all my lectures and}  \textbf{tutorials for me to night classes at the}  \textbf{exact same times}! \\ 
 R5   & Return Gerund and Infinitive verb phrases               & \textbf{Being covered in hives} is so much fun!; Can't wait \textbf{to wake up early to babysit}...                                                           \\ 
 R6   & Return nouns preceded by a positive sentiment adjective            & Yep, this is indeed an \textit{amazing} \textbf{donut} ..      \\ 
 R7   & Return subject of interrogative sentences  & A murderer is stalking me. Could \textbf{life} be  more fun?                                                                                     \\ 
 R8   & Return subjects of  comparisons (similes)      & \textbf{He} is as good at coding as \textbf{Tiger Woods} is at avoiding controversy.                                                                         \\ 
 R9   & Return demonstrative adjective-noun pairs & Oh, I love \textbf{this jacket}!  \\ \bottomrule
 \end{tabular}
 \caption{Summary of rules in the rule-based extractor}
 \label{tab:rules}
 \end{table*}
\textbf{Combining the outputs of individual rules to generate candidate sarcasm targets of the rule-based extractor}: To generate the set of candidate sarcasm targets returned by the rule-based extractor, a weighted majority approach is used as follows. Every rule above is applied to the input sarcastic sentence. Then, every word is assigned a score that sums the accuracy of rules which predicted that this word is a part of the sarcasm target. This accuracy is the overall accuracy of the rule as determined by solely the rule-based classifier\footnote{These values are shown in Tables 4 and 5}. Thus, the integrator weights each word on the basis of how good a rule predicting it as a target was. Words corresponding to the maximum value of this score are returned as candidate sarcasm targets.

\subsection{Statistical Extractor}
The statistical extractor uses a classifier that takes as input a word (along with its features) and returns if the word is a sarcasm target. To do this, we decompose the task into $n$ classification tasks, where $n$ is the total number of words in the sentence. This means that every word in input text is considered as an instance, such that the label can be 1 or 0 depending on whether or not the given word is a part of sarcasm target. For example, `\textit{Tooth-ache is fun}' with sarcasm target as `\textit{tooth-ache}' is broken down into three instances: `\textit{tooth-ache}' with label 1, `\textit{is}' with label 0 and `\textit{fun}' with label 0. In case the target lies outside the sentence, all words have the label 0. 

We then represent the instance (\textit{i.e.}, the word) as a set of following features: (A) \textbf{Lexical}: \textit{Unigrams}, (B) \textbf{Part of Speech (POS)}-based features: \textit{Current POS}, \textit{Previous POS}, \textit{Next POS},
(C) \textbf{Polarity}-based features: \textit{Word Polarity} : Sentiment score of the word, \textit{Phrase Polarity} : Sentiment score for the trigram formed by considering the previous word, current word and the next word together (in that order). These polarities lie in the range [-1,+1]. These features are based on our analysis that the target phrase or word tends to be more neutral than the rest of the sentence, and
(D) \textbf{Pragmatic features}: \textit{Capitalization} : Number of capital letters in the word. Capitalization features are chosen based on features from ~\newcite{4}.

% \begin{table}[]
% \centering

% \begin{tabular}{ll}
% \toprule
% \textbf{Feature} & \textbf{Description} \\ \midrule
% \multicolumn{2}{c}{\textbf{Lexical}}                                               \\ \midrule
% Unigram         & Words                                   \\ \midrule
% \multicolumn{2}{c}{\textbf{Part of Speech}}                                        \\ \midrule
% Current POS     & POS tag of the current word                                        \\ 
% Previous POS    & POS tag of the previous word                                       \\
% Next POS        & POS tag of the next word                                           \\ \midrule
% \multicolumn{2}{c}{\textbf{Sentimental Polarity}}                                  \\ \midrule
% Word Polarity   & Polarity of the current word                                       \\ 
% Phrase Polarity & Polarity of the phrase formed by "previous + current + next" words \\ \midrule
% \multicolumn{2}{c}{\textbf{Pragmatic}}                                             \\ \midrule
% Capitalization  & Number of capital letters in the word.                             \\ \bottomrule
% \end{tabular}
% \caption{Features of our Statistical Extractor}
% \label{features}
% \end{table}

The classifiers are trained with words as instances while the sarcasm target is to be computed at the sentence level. Hence, the candidate sarcasm target returned by the statistical extractor consists of words for which the classifier returned 1. For example, the sentence `This is fun' is broken up into three instances: `this', `is' and `fun'. If the classifier returns 1, 0, 0 for the three instances respectively, the statistical extractor returns `this' as the candidate sarcasm target. Similarly, if the classifier returns 0, 0, 0 for the three instances, the extractor returns the fallback label `Outside'.
%For each word in a particular sentence, our classifier makes the prediction, and the actual target list is compared with the list of words predicted to be target by the classifier for overlap. Correct overlaps are recorded for calculating accuracy (results are discussed in the Evaluation section).
%\subsubsection{Details of the SVM classifier}

\subsection{Integrator}
The integrator determines the sarcasm target based on the outputs of the two extractors. We consider two configurations of the integrator:
\begin{enumerate}\setlength\itemsep{0cm}
\item Hybrid \textbf{OR}: In this configuration, the integrator predicts the set of words that occur in the output of either of the two extractors as the sarcasm target. If the lists are empty, the output is returned as `Outside'.
\item Hybrid \textbf{AND} : In this configuration, the integrator predicts the set of words that occur in the output of both the two extractors as the sarcasm target. If the intersection of the lists is empty, the output is returned as `Outside'.
\end{enumerate}
The idea of using two configurations OR and AND is based on a rule-based sarcasm detector by
\cite{precedes}. While AND is intuitive, the second configuration OR is necessary because our extractors individually may not capture all forms of sarcasm target. This is intuitive because \textit{our rules may not cover all forms of sarcasm targets}.
%\subsection{Sequence Labeling}
%This approach treats every instance (sentence) under consideration as a connected sequence of words i.e treating words as \textit{an ordered sequence} and not a collection of discrete units as done in statistical approach. We use the same features as the statistical approach.\\
%This time, we use the \textbf{SVM:HMM tool}, again with \textbf{four fold cross validation}.

\begin{table}[]
\centering
\small
\begin{tabular}{p{3.9cm}ll}
 \toprule                                    & \textbf{Snippets} & \textbf{Tweets} \\ \midrule
Count                                                          & 224           & 506    \\
Average \#words                                                & 28.47         & 13.06  \\
Vocabulary                                                     & 1710          & 1458   \\
Total words & 6377 & 6610 \\
Average length of sarcasm target                               & 1.6           & 2.08   \\
Average polarity strength of sarcasm target                    & 0.0087        & 0.035  \\
Average polarity strength of portion apart from sarcasm target & 0.027         & 0.53  \\ \bottomrule
\end{tabular}
\caption{Statistics of our datasets; `Snippets': Book Snippets}
\label{tab:stats}
\end{table}
\section{Experiment setup}
\label{sec:expsetup}
We evaluate our approach using two datasets: one consisting of book snippets and another of tweets. The dataset of book snippets is a sarcasm-labeled dataset by ~\newcite{adityaemnlp}. 224 book snippets marked as sarcastic are used. On the other hand, for our dataset of tweets, we use the sarcasm-labeled dataset by ~\newcite{riloff}. 506 sarcastic tweets from this dataset are used. The statistics of the two datasets are shown in Table~\ref{tab:stats}. The average length of a sarcasm target is 1.6 words in case of book snippets and 2.08 words in case of tweets. The last two rows in the table point to an interesting observation. In both the datasets, the average polarity strength\footnote{Polarity strength is the sum of polarities of words. We use a sentiment word-list to get the strength values} of sarcasm target is lower than polarity strength of rest of the sentence. This shows that sarcasm target is likely to be more neutral than sentiment-bearing. Note that all textual units (tweets as well as book snippets) in both datasets are sarcastic.

We use \textbf{SVM Perf}~\cite{joachims2006training} to train the classifiers, optimized for F-score with epsilon e=0.5 and RBF kernel\footnote{RBF Kernel performed better than linear kernel.}. We set C=1000 for tweets and C=1500 for snippets. We report our results on \textbf{four-fold cross validation} for both datasets. \underline{Note} that we convert individual sentences into words. Therefore, the dataset in case of book snippets has 6377 instances, while the one of tweets has 6610 instances. The four folds for cross-validation are created over these instances. With a word as instance, the task is binary classification: 1 indicating that the word is a sarcasm target and 0 indicating that it is not. For rules in the rule-based extractor, we use tools in NLTK~\cite{nltk}, wherever necessary. 

We consider two baselines with which our hybrid approach is compared:
\begin{enumerate}\setlength\itemsep{0cm}
\item \textbf{Baseline 1: All Objective Words}:
As the first baseline, we design a na\"{i}ve approach for our task: include all words of the sentence which are not stop words, and have neutral sentiment polarity, as the predicted sarcasm target. We reiterate that in case of papers with no past work, simplistic baselines have been commonly reported in sentiment analysis. However, to validate that our hybrid approach is valuable, we compare our hybrid approach against other possible versions of the system as well.
\item \textbf{Baseline 2}: Baseline 2 is derived from past work in opinion target identification because sarcasm target identification may be considered as but a form of opinion target identification. Sequence labeling has been reported for opinion target identification~\cite{jin2009opinionminer}. Therefore, we use SVM-HMM~\cite{svmhmm} with default parameters as the second baseline.
\end{enumerate}

We report performance using two metrics: Exact Match Accuracy and Dice Score. These metrics have been used in past work in information extraction~\cite{michelson2007unsupervised}. As per their conventional use, these metrics are computed at the sentence level. The metrics that we use are:
\begin{itemize}\setlength\itemsep{0cm}
%\item \textbf{Partial Match Accuracy} : A partial match occurs if there is at least one element (word) common to the predicted sarcasm target list and the actual target list. The accuracy is computed as number of instances with partial match divided by total instances.
\item \textbf{Exact Match (EM) Accuracy} : An exact match occurs if the list of predicted target(s) is exactly the same as the list of actual target(s). The accuracy is computed as number of instances with exact match divided by total instances.%The total number of instances (sentences) with a partial (exact) match is divided by the size of the dataset to give the final result for partial (exact) match accuracy.
\item \textbf{Dice Score} : Dice score\cite{sorensen1948method} is used to compare similarity between two samples. This is considered to be a better metric than Exact match accuracy because it accounts for missing words and extra words in the target.%Let the two lists (predicted and actual) be X and Y. Dice score is given by \textbf{\(( 2X\cap Y )/( X+Y )\)}.
\end{itemize}
\begin{table}[h]
\centering
\begin{tabular}{lcccc}
\toprule
\multicolumn{1}{c}{\multirow{2}{*}{\textbf{Rule}}} & \multicolumn{2}{c}{\textbf{Overall}}                              & \multicolumn{2}{c}{\textbf{Conditional}}                                                                                                                                                         \\ %\cline{2-7} 
\multicolumn{1}{c}{}                               &  \textbf{EM} & \textbf{DS}& \textbf{EM} & \textbf{DS} \\ \midrule %
\textbf{R1}                                                               & 7.14                                                           & \textbf{32.8}                                                                    & 7.65                                                           & 35.23                                                         \\ 
\textbf{R2}                                                          & \textbf{8.48}                                                  & 16.7                                                                      & 19.19                                                          & 37.81                                                         \\ 
\textbf{R3}              & 4.91                                                           & 6.27                                                                                                                 & 16.92                                                          & 21.62                                                         \\ 
\textbf{R4}                                                                             & 2.67                                                           & 11.89                                                                                                      & 4.38                                                           & 19.45                                                         \\ 
\textbf{R5}                                                                       & 1.34                                                           & 6.39                                                                                                               & 2.32                                                           & 11.11                                                         \\ 
\textbf{R6}                                                               & 4.01                                                           & 6.77                                                                          & 8.91                                                           & 15.02                                                         \\ 
\textbf{R7}                                                                    & 3.12                                                           & 10.76                                                                                                      & 9.46                                                           & 32.6                                                          \\ 
\textbf{R8}                                                           & 4.91                                                           & 6.78                                                                              & \textbf{35.02}                                                 & 45.17                                                         \\ 
\textbf{R9}                                                                      & 4.46                                                           & 6.94                                                                                                    & 34.48                                                          & \textbf{53.67}                                                \\ \bottomrule
\end{tabular}
\caption{Results for individual rules for book snippets}
\label{rule_results_snippets}
\end{table}

\begin{table}[h]
\centering
\begin{tabular}{lcccc}
\toprule
\multicolumn{1}{c}{\multirow{2}{*}{\textbf{Rule}}} & \multicolumn{2}{c}{\textbf{Overall}}                                                                                                                                                              & \multicolumn{2}{c}{\textbf{Conditional}}                                                                                                                                                         \\  %\cline{2-7} 
 \multicolumn{1}{c}{}                               & \textbf{EM} & \textbf{DS} & \textbf{EM} & \textbf{DS} \\ \midrule %
\textbf{R1}     & 6.32                                                           & 19.19                                                        & 8.69                                                           & 26.39                                                         \\ 
\textbf{R2}       & 11.26                                                          & 16.18                                                          & 30.32                                                          & 43.56                                                         \\ 
\textbf{R3}                                                                  & \textbf{12.45}                                                 & 20.28                                                                                                       & 34.24                                                          & \textbf{55.77}                                                \\ 
\textbf{R4}                                                             & 6.91                                                           & 13.51                                                                                                                & 18.42                                                          & 36.0                                                          \\ 
\textbf{R5}                & 9.28                                                           & \textbf{23.87}                                                  & 15.36                                                          & 39.47                                                         \\ 
\textbf{R6}                           & 10.08                                                          & 16.91                                                          & 19.31                                                          & 32.42                                                         \\ 
\textbf{R7}                             & 9.88                                                           & 15.21                                                                   & 32.25                                                          & 49.65                                                         \\ 
\textbf{R8}                                                                     & 11.26                                                          & 11.26                                                           & \textbf{50}                                                    & 50                                                            \\ 
\textbf{R9}                                                           & 11.46                                                          & 13.28                                                                                                             & 43.59                                                          & 50.51                                                         \\ \bottomrule
\end{tabular}
\caption{Results for individual rules for tweets}
\label{rules_results_tweets}
\end{table}
%\textbf{Motivation behind use of dice score} : As we can see, the partial match is a bit too lenient, while the exact match is extremely tight for evaluating this new concept. Dice score can effectively capture the similarity between our predictions and actual target, and serves as a more reliable measure of the strength of the system under consideration (the approach, or the individual rule).
    
\textbf{Note}: If the actual target is the fallback label `Outside', then the expected predicted target is either `Outside' or empty prediction list. In such a case, the instance will contribute to exact match accuracy.
\section{Results}
\label{sec:results}
This section presents our results in two steps: performance of individual rules that are a part of the rule-based extractor, and performance of the overall approach.
\subsection{Performance of rules in the rule-based extractor}
Tables ~\ref{rule_results_snippets} and ~\ref{rules_results_tweets} present the performance of the rules in our rule-based extractor, for snippets and tweets respectively. The two metrics (exact match accuracy and dice score) are reported for two cases: Overall and Conditional. `Overall' spans all text units in the dataset whereas `Conditional' is limited to text units which match a given rule (\textit{i.e.}, where the given linguistic phenomenon of, say, gerunds, etc. is observed). Considering the `Conditional' case is crucial because a rule may be applicable for a specific form of sarcasm target, but may work accurately in those cases. Such a rule will have a low `overall exact match/dice score' but a high `conditional exact match/dice score.' Values in bold indicate the best performing rule for a given performance metric. As seen in the tables, the values for `conditional' are higher than those for `Overall'. For example, consider rule \textbf{R7} in Table~\ref{rule_results_snippets}. Exact match of 3.12 (for overall accuracy) as against 9.46 (for conditional accuracy). This situation is typical of rule-based systems where rules may not cover all cases but be accurate for situations that they do cover. %This is likely to be because R5 returns gerundial phrases and infinitives as target. Such phrases may occur infrequently in sarcastic text, however, when they do, they are likely to be sarcasm targets. For five out of nine rules, the increase in the metric values is high.  This also holds for rules R3, R4, R8 and R9. This shows that these rules may be specific (\textit{i.e.}, apply to a small set of examples) but are \textit{powerful} (\textit{i.e.}, have a high conditional accuracy). This scenario of coverage of rules is obvious in case of rule-based systems.

For tweets, R3 has a very high dice score (conditional) (55.77). This rule validates the benefit of utilizing the structure of sarcastic tweets as explored by \cite{riloff} : `contrast of positive sentiment with negative situation' being a strong indicator of sarcasm target. %Sarcastic tweets tend to be strongly positive on surface (with positive sentiment verb like 'love' quite 'common'), but criticize or mock the target.
% For both the datasets, we can see that rule 8 (R8) (in conditional case) shows good results for exact match. This is likely to occur in situations like similes. 

\subsection{Overall Performance}
We now compare the performance of the approach with the baseline (as described in Section~\ref{sec:expsetup}). In order to understand the benefit of individual extractors, we also show their performance when they are used individually. Thus, we compare five approaches: (A) Baseline, (B) Rule-based (when only the rule-based extractor is used), (C) Statistical (when only the statistical extractor is used), and (D) \& (E) Hybrid (two configurations: OR and AND). It must be noted that \textbf{since no existing sarcasm target identification approach exists, we rely on the approach of using a simple baseline, and verify if our approach does any better than a simpler, obvious baseline}. Such baselines have been used in early work in sentiment analysis. For example, ~\newcite{pang2005seeing} compare against a `random-choice' baseline, or ~\newcite{tan2011user} who use a simple majority-voting baseline, in absence of past work. We also use a second baseline from a related area: sentiment/opinion target identification.
%We will first present our results for all the various approaches we use, with one table each for both of our datasets : snippets and Tweets. Then we will discuss the results and our observations.

Tables~\ref{results_snippets_1} and ~\ref{results_tweets_1} compare the five approaches for snippets and tweets respectively. All our approaches outperform the baseline in case of exact match and dice score. In case of tweets, Table ~\ref{results_tweets_1} shows that the rule-based extractor achieves a dice score of 29.13 while that for statistical extractor is 31.8. Combining the two together (owing to our hybrid architecture) improves the dice score to 39.63. This improvement also holds for book snippets. This \textbf{justifies the `hybrid' nature of our approach}. Hybrid OR performs the best in terms of Dice Score. However, for exact match accuracy, Hybrid AND achieves the best performance (16.51 for snippets and 13.45 for tweets). This is likely because Hybrid AND is restrictive with respect to the predictions it makes for individual words. The statistical extractor performs better than rule-based extractor for all three metrics. For example, in case of tweets, the dice score for statistical extractor is 31.8 while that for rule-based extractor is 29.13. Also, nearly all results (across approaches and metrics) are higher in case of tweets as compared to snippets. Since tweets are shorter than snippets (as shown in Table~\ref{tab:stats}), it is likely that they are more direct in their ridicule as compared to snippets.

%This is likely caused by the fact that this is a strict approach (word is the target if and only if it is predicted to be so by both SVM and rule based classifier), and exact match is a very strict metric in itself.

%In both the tables, the first two rows show that the statistical approach produces better results than the rule based approach (for all three metrics).

%In table 3 (snippets), we can see that sequence labeling approach really gives the same results as statistical approach, and very small improvements in case of table 4. This shows that the idea of exploiting the 'sequential order of words in the sentence' through sequence labeling does not really help with sarcasm target identification.

%We can see that the baseline (the last row in both the tables) has a very high accuracy in case of partial match. By definition, our baseline approach captures a large part of the sentence as the predicted target. Due to this, the chances of a partial match are very high, but the same reason causes a very low (almost negligible) accuracy for exact match. As the dice score indicates, the four approaches we used (rule-based,statistical,hybrid, and sequence labeling) are clearly more powerful or accurate for capturing the sarcasm target as compared to the baseline.

%It is interesting that  In contrast, snippets dataset often has the snippets from a conversation, where people are likely to be more vague from a stranger's (annotator) perspective.

%Thus the short and direct nature of the tweets helps in better sarcasm target identification.

\begin{table}[h]
\centering
\begin{tabular}{p{4.5cm}cc}
\toprule
 \textbf{Approach}    & \textbf{EM} & \textbf{DS} \\ \midrule
\textbf{Baseline 1: All Objective Words}           &  0.0                  & 16.14               \\
\textbf{Baseline 2: Seq. Labeling}           & 12.05                  & 31.44               \\
\textbf{Only Rule-Based}         & 9.82                 & 26.02               \\ 
\textbf{Only Statistical}      & 12.05                & 31.2                \\ 
\textbf{Hybrid OR}  & 7.01                 & \textbf{32.68}               \\ 
\textbf{Hybrid AND}    & \textbf{16.51}               & 21.28               \\ 
%\multicolumn{2}{l}{\textbf{Sequence Labeling}} & 50                     & 12.05                & 31.44               \\ 
 \bottomrule
\end{tabular}
\caption{Performance of sarcasm target identification for snippets}
\label{results_snippets_1}
\end{table}

\begin{table}[h]
\centering
\begin{tabular}{p{4.5cm}cc}
\toprule
\textbf{Approach}     &  \textbf{EM} & \textbf{DS} \\ \midrule
\textbf{Baseline 1: All Objective Words}            & 1.38                 & 27.16               \\ 
\textbf{Baseline 2: Seq. Labeling}   & 12.26                  & 33.41               \\
\textbf{Only Rule-Based}  & 9.48                 & 29.13               \\ 
\textbf{Only Statistical}     & 10.48                & 31.8                \\ 
\textbf{Hybrid OR}  & 9.09                 & \textbf{39.63}               \\ 
\textbf{Hybrid AND} & \textbf{13.45}                & 20.82               \\ 
%\multicolumn{2}{l}{\textbf{Sequence Labeling}} & 53.96                  & 12.26                & 33.41               \\ 
\bottomrule
\end{tabular}
\caption{Performance of sarcasm target identification for tweets}
\label{results_tweets_1}
\end{table}

%\section{Discussion}

% Why is this discussion necessary? (Two sentences)
% Presence of outside cases: statistics and 3 examples (4 sentences)
% Paragraph change
% Performance of Hybrid OR for outside: 1 sentence - what results you have, 2 sentences - (a) Read the value, (b) Compare it with overall results.
% Upar ke 3 examples mein se kitne sahi, kitne galat aaye - describe that, giving possible reasons. (2 sentences)
% Error analysis: Only example and heading: (3 examples)

%A peculiar challenge for sarcasm target identification is that a sarcasm target may lie outside the text. Hence, in this section, we discuss the performance of our approach for these cases (we refer to such cases as `outside' cases). In our dataset of book snippets, there are 11 texts (~5\%) with sarcasm target outside the text. In case of tweets, such cases are much higher: 53 tweets (~10\%). 

%In case of the three examples above, example 1 (a tweet) is correctly identified, while example 2 (snippet) and example 3 (tweet) are not identified to be the `outside' case. For the `apples' example, our system predicted `I' as the sarcasm target. For the `senior year' example, the (incorrectly) predicted target is `me'.
\begin{table}[h]
\centering
\small
\begin{tabular}{p{1.9cm}llll}
\toprule
                    & \multicolumn{2}{c}{\textbf{Book Snippets}}       & \multicolumn{2}{c}{\textbf{Tweets}  }            \\ \midrule
                     & \textbf{EM} & \textbf{DS}  &\textbf{EM} & \textbf{DS}  \\ 
Overall                & 7.01        & 32.68          & 9.09        & 39.63      \\ 
`Outside' cases      & 6.81        & 6.81              & 4.71        & 4.71       \\ \bottomrule
\end{tabular}
\caption{Comparison of performance of our approach in case of examples with target outside the text (indicated by `Outside' cases), with complete dataset (indicated by `Overall'); EM: Exact Match, DS: Dice Score}
\label{tab:discuss}
\end{table}
 \section{Error Analysis}
 \label{sec:discuss}
 A key source of error is cases where the target lies outside the text. In this section, we describe such examples and compare the impact of these errors with the overall performance.
In our dataset of book snippets, there are 11 texts (~5\%) with sarcasm target outside the text. In case of tweets, such cases are much higher: 53 tweets (~10\%). Table ~\ref{tab:discuss} compares the results of our hybrid (OR) approach for the specific case of target being `outside' the text (indicated by `Outside cases' in the table), with the results on the complete dataset (indicated by `Overall' in the table). Dice Score (DS) for book snippets is 6.81 for `outside' cases as compared to 32.68 for the complete dataset. In general, the performance for the `outside' cases is  lower than the overall performance. This proves the difficulty that the `Outside' cases presents. The EM and DS values for `Outside' cases are the same by definition. This is because when the target is `Outside', a partial match and an exact match are the same. Our approach correctly predicts the label `Outside' for sentences like `\textit{Yeah, just ignore me. That is TOTALLY the right way to handle this!}' However, our approach gives the incorrect output for some examples. For example, for `\textit{Oh, and I suppose the apples ate the cheese}', the predicted target is not `Outside' (the expected label) but `\textit{I}'. Similarly, for `\textit{Please keep ignoring me for all of senior year. It's not like we're friends with the exact same people}', the incorrectly predicted target is `\textit{me}' instead of the expected label `Outside'.

\section{Related Work}
\label{sec:relwork}
Computational sarcasm primarily focuses on sarcasm detection: classification of a text as sarcastic or non-sarcastic. \newcite{joshi2016automatic} present a survey of sarcasm detection approaches. They observe three trends in sarcasm detection: semi-supervised extraction of sarcastic patterns, use of hashtag-based supervision, and use of contextual information for sarcasm detection~\cite{3,4,22}. However, to the best of our knowledge, no past work aims to identify phrases in a sarcastic sentence that indicate the target of ridicule in the sentence. 

Related to sarcasm target identification is sentiment target identification. Sentiment target identification deals with identifying the entity towards which sentiment is expressed in a sentence. \newcite{qiu2011opinion} present an approach to extract opinion words and targets collectively from a dataset. Aspect identification for sentiment has also been studied. This deals with extracting aspects of an entity (for example, color, weight, battery in case of a cell phone). Probabilistic topic models have been commonly used for the same. \newcite{titov2008joint} present a probabilistic topic model that jointly estimates sentiment and aspect in order to achieve sentiment summarization. \newcite{lu2011multi} perform multi-aspect sentiment analysis using a topic model. Several other topic model-based approaches to aspect extraction have been reported~\cite{mukherjee2012aspect}. To the best of our knowledge, ours is the first work that deals with sarcasm target identification.

\section{Conclusion \& Future Work}
\label{sec:concl}
In this paper, we introduced a novel problem: sarcasm target identification. This problem aims to identify the target of ridicule in a sarcastic text. This target may be a subset of words in the text or a fallback label `Outside'. The task poses challenges such as multiple sarcasm targets or sarcasm targets that may not even be present as words in the sentence. We present an introductory approach for sarcasm target identification that consists of two kinds of extractors: a rule-based and a statistical extractor. Our rule-based extractor implements nine rules that capture forms of sarcasm target. The statistical extractor splits a sentence of length $n$ into $n$ instances, where each instance is represented by a word, and a label that indicates if this word is a sarcasm target. A statistical classifier that uses features based on POS and sentiment, predicts if a given word is likely to be a target or not. Finally, an integrator combines the outputs of the two extractors in two configurations: OR and AND. We evaluate our approach on two datasets: one consisting of snippets from books, and another of tweets. In general, our hybrid OR system performs the best with a Dice score of 39.63. This is higher than two baselines: a na\"{i}ve baseline designed for the task, and a baseline based on sentiment target identification. Our hybrid approach is also higher than the two extractors individually used. This shows that the two extractors collectively form a good sarcasm target identification approach. Finally, we discuss performance in case of examples where the target is outside the sentence. In such cases, our approach performs close to the overall system in terms of exact match, but there is a severe degradation in Dice score. We finally present an analysis of errors due to target being outside the text.

Our work forms a foundation for future approaches to identify sarcasm targets. As future work, additional rules in the rule-based extractor and novel sets of features in the statistical extractor may be used. Use of syntactic dependencies has been found to be useful in case of opinion target extraction~\cite{qiu2011opinion}. Applying these techniques for sarcasm target identification can be useful. A special focus on the `outside' cases (\textit{i.e.}, cases where the target of ridicule in a sarcastic text is beyond the words present in the sentence) is likely to be helpful for sarcasm target identification. 

\bibliographystyle{acl2012}
\bibliography{main}

\end{document}